\documentclass[runningheads]{llncs}
\usepackage[T1]{fontenc}
\usepackage[utf8]{inputenc}
\usepackage[small]{caption}
\usepackage{graphicx}
\usepackage{amsmath}

\usepackage{amsthm}
\usepackage{booktabs}
\usepackage{algorithm}
\usepackage{algorithmic}
\usepackage[switch]{lineno}
\usepackage{amssymb}
\usepackage{makecell}
\usepackage{multirow}
\usepackage{color}
\UseRawInputEncoding

\begin{document}
\title{Detect an Object At Once without Fine-tuning}
%
%


\author{Junyu Hao\inst{1,2*} \and
	Jianheng Liu\inst{3*} \and
	Yongjia Zhao\inst{4}\and
    Zuofan Chen\inst{5}\and
    Qi Sun\inst{5}\and    
	Jinlong Chen\inst{5}\and
	Jianguo Wei\inst{1} \and
	Minghao Yang\inst{2}$^{\#}$
   }

\authorrunning{F. Author et al.}
%
\institute{School of Computer Software, Tianjin University, Tianjin, China \and
Institute of Automation, Chinese Academy of Sciences, Beijing, China \and
University College London, London, United Kingdom \and
Beihang University, Beijing, China \and	
School of Information Engineering, Zhejiang Sci-Tech University, Hangzhou, China \and
	School of computer science and information security, Guilin university of electronic Technology, Guilin, China\\	
	\email{haojy@tju.edu.cn,jianheng.liu.23@ucl.ac.uk,zhaoyongjia@buaa.edu.cn,
		\\2023210601001@mails.zstu.edu.cn,sunqi@zstu.edu.cn,jinlong.chen@guet.edu.cn,\\jianguo@tju.edu.cn,mhyang@nlpr.ia.ac.cn}}

\maketitle              
\begin{abstract}
When presented with one or a few photos of a previously unseen object, humans can instantly recognize it in different scenes. Although the human brain mechanism behind this phenomenon is still not fully understood, this work introduces a novel technical realization of this task. It consists of two phases: (1) generating a Similarity Density Map (SDM) by convolving the scene image with the given object image patch(es) so that the highlight areas in the SDM indicate the possible locations; (2) obtaining the object occupied areas in the scene through a Region Alignment Network (RAN). The RAN is constructed on a backbone of Deep Siamese Network (DSN), and different from the traditional DSNs, it aims to obtain the object accurate regions by regressing the location and area differences between the ground truths and the predicted ones indicated by the highlight areas in SDM. By pre-learning from labels annotated in traditional datasets, the SDM-RAN can detect previously unknown objects without fine-tuning. Experiments were conducted on the MS COCO, PASCAL VOC datasets. The results indicate that the proposed method outperforms state-of-the-art methods on the same task.

\end{abstract}

\section{Introduction}
Given one or a couple of photos of an object unseen before, machines are expected to detect it quickly and accurately in various scenes. This ability is advantageous in a variety of applications, such as quickly detecting unfamiliar objects \cite{DiGeo2023,ARPN2020,Laura2021,Yang2020}, autonomous exploration in unknown environments \cite{AirDet2022}, rapid count for density objects scenes \cite{Viresh2021,Carlos2016}, etc. Although humans have this ability, the underlying brain mechanism is still unclear. This task has motivated the recent development of few-shot object detection at once (FSOD-AO). A practical FSOD-AO method demands identifying the categories of the new objects correctly in scenes, meanwhile extracting their accurate regions simultaneously without fine-tuning.

Two prior works on FSOD-AO have recently been proposed by researchers: Attention-Region Proposal Network (A-RPN) \cite{ARPN2020} and AirDet \cite{AirDet2022}. These two works considered FSOD-AO as a region proposal task between the few shot support images and query images, while suppressing false detections in the background. In addition, AirDet paid attention to retrieving the multi-scale features from the cross-scale relationships between the support and query images. However, A-RPN and AirDet used the Region Proposal Network (RPN) to generate the candidates and regions, and did not consider an accurate box matching and alignment mechanism.

This work proposes a novel FSOD-AO method. Unlike A-RPN and AirDet, where RPN is used to find the candidates, this work obtains the candidate positions using the Similarity Density Map (SDM) (inspired by the idea presented in the article \cite{Viresh2021}). Furthermore, we introduce a Region Alignment Network (RAN) to suppress false detections and align the candidate positions synchronously. The proposed method (SDM-RAN) is run and estimated on two widely recognized datasets (MS COCO and PASCAL VOC). The results show that the proposed method outperforms A-RPN and AirDet in novel object detection without fine-tuning.


\section{Related Work}

\subsection{Few-shot Object Detection}

Few-Shot Object Detection (FSOD) is different from mainstream deep architecture detection methods in its goal to detect new objects by using either one or a few annotated images. Meanwhile, unlike early methods that relied on small sample sizes and object retrieval techniques, such as category-related template \cite{Pedro2005}, point distributed model  \cite{Andriy2010}, or shape-invariant strategie \cite{WeiLian2012}, etc., FSOD can detect targets without the need for tedious feature or template selection.

There are three main categories of fine-tuning FSOD methods: transfer learning, metric-learning, and meta-learning. Transfer learning approaches refer to the reuse of network weights pre-trained on a source domain to improve generalization capabilities on a few novel images \cite{Aming2021,MPSR2020}. Recent work has introduced semantic relations between novel base classes \cite{Chenchen2021} and contrastive proposals \cite{Sunbo2021}. However, transfer learning is highly dependent on the source domain and is difficult to extend to very different scenarios.

Metric-learning aims to learn the appropriate strategies, or embedding spaces, where similar content is encoded in features with small distances to each other, while the encoded features of dissimilar inputs should be far apart \cite{Jake2017,Boris2018,Bharath2017}. Since the training samples are constructed in pairs, triplets or quadruplets, there is a polynomial growth of training pairs, which are highly redundant and less informative \cite{Xun2019}. If the novel objects are similar to certain objects in the source domain, the metric-learning model usually needs to be re-trained several times before possible acceptable performance is achieved \cite{Oriol2016,Eleni2017}.

Meta-learning learns generalization strategies for new tasks or new data \cite{Timothy2022}. They need to learn how to learn the necessary category knowledge efficiently, so that the knowledge can be applied to novel categories with few training examples. A more recent trend in meta-learning is to design general agents that can guide supervised learning within multiple tasks by accumulating knowledge from training examples \cite{Tsendsuren2018,Finn2017,Kwonjoon2019}. However, traditional meta-learning methods still lack mechanisms to associate the support information and the queried image for category identification and region alignment. Meta-learning has mostly focused on the classification tasks, but rarely on the accurate object detection from few shots.

\subsection{Few-shot Object Detection At Once (FSOD-AO)}
Unlike FSOD, FSOD-AO requires machines to identify a novel, previously unseen object in different scenes without fine-tuning. FSOD-AO has three core requirements: finding the candidate regions of target objects, retaining the correct categories by suppressing the false ones, and accurately identifying the target regions. These steps are required for processing without fine-tuning. Attention-Region Proposal Network (A-RPN) \cite{ARPN2020} is an early pioneering method proposed for the FSOD-AO task. It uses the region proposal network (RPN) originally proposed in Faster-RCNN \cite{FasterRCNN2015} to find the class-agnostic relational regions, and then uses a multi-relational classifier to suppress errors. Another pioneering work, AirDet \cite{AirDet2022}, also uses RPN to obtain possible regions, and then uses Global-Local Relation (GLR) and Relation Embedding (PRE) for location regression. However, both methods lack an accurate mechanism for refining regions, as we introduced in the first section.

A related work to FSOD-AO is visual counting. It aims to obtain the overall number of a particular category at a time, even if the novel objects are very dense in the scene \cite{Carlos2016}. FamNet \cite{Viresh2021} takes multiple objects from the query image and predicts a density map for the presence of all objects of interest in the query image. Although FamNet can quickly count the novel visual category by estimating the number of highlight points in the density map, it only presents the center points of the targets and ignores the occupied areas of the targets in the scene.

Given an object patch and its position in the previous video frame, Deep Siamese Networks (DSN) can determine its new position in subsequent frames \cite{Luca2016,Boli2016}. The proposed FSOD-AO method is partly inspired by the idea of FamNet and DSN, namely we consider the detection of an unseen object as two phases: exploring the possible locations of new objects in the scenes, and immediately determining their regions using DSN.

\begin{figure*}
	\begin{center}
		\includegraphics[width=1 \linewidth]{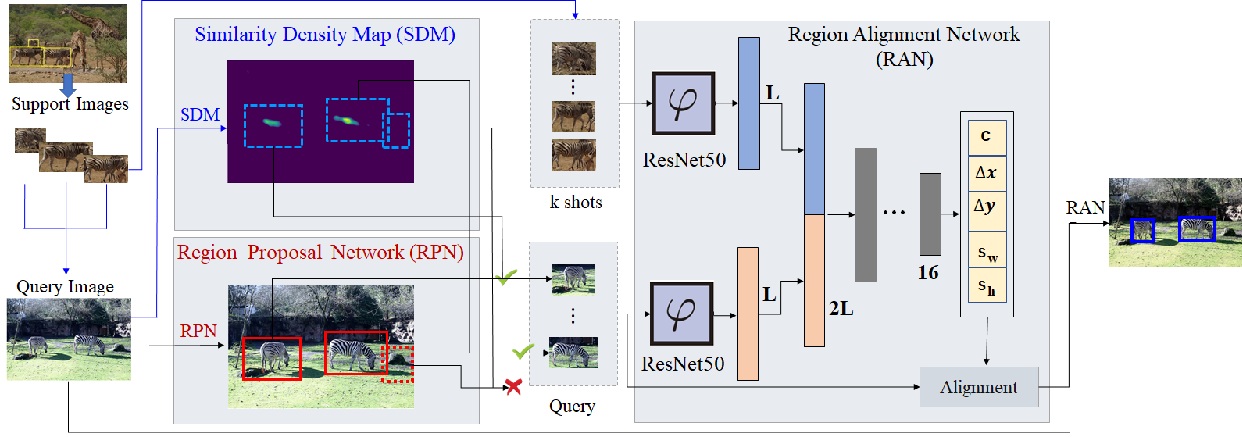}
	\end{center}
	\caption{The framework of the proposed method.}
	\label{fig:frame_work}
\end{figure*}

\section{Methodology}
\subsection{Framework}
\subsubsection{Definition}
Before introducing the proposed method, we first give the preliminary concepts. In this work, all targets are divided into two classes: $C_{B}$ and $C_{F}$ ($C_{B} \cap C_{F}$=$\varnothing$), where $C_{B}$ is the bases classes, and $C_{F}$ is the few shot detection set containing $N$ novel categories. During the training process, images from the base classes $C_{B}$ are split into query images $C_{B}^{q}$ and support images $C_{B}^{s}$ ($C_{B}$=$C_{B}^{q}$+$C_{B}^{s}$). Given all support images $C_{B}^{s}$, the model learns to detect objects in query images $C_{B}^{q}$. In the test phase, images from the few shot detection set $C_{F}$ are split into query images $C_{F}^{q}$ and support images $C_{F}^{s}$ ($C_{F}$=$C_{F}^{q}$+$C_{F}^{s}$). The model detects objects in unknown query images $C_{F}^{q}$ by only providing $k$ ($1\leq k\leq K$) support images in $C_{F}^{s}$ without fine-tuning (the values of $K$ are usually in the range of 1-5) \cite{AirDet2022}.

\subsubsection{Workflow}
Fig.~\ref{fig:frame_work} outlines the framework of the proposed FSOD-AO method. It mainly consists of three modules: Location Prediction using Similarity Density Map (SDM), Region Proposal Network (RPN), Region Alignment Network (RAN). SDM is  presented for the images from $C_{B}^{s}$, and it can immediately obtain objects’ possible centers in $C_{B}^{q}$ with dot annotations \cite{Viresh2021}. RPN is previously trained with region-annotations. It aims to obtain objects’ possible regions in $C_{B}^{q}$. With the results obtained by SDM and RPN, they are followed by RAN to suppress the false detections in the background, and also to adjust the predicted box region as well.

\vspace{-0.3cm} 
\subsection{Location Prediction (SDM)}
Inspired by \cite{Viresh2021}, we use the Few Shot Adaptation and Matching Network (FamNet) to obtain objects’ possible center locations. FamNet mainly contains a feature extraction module and a density prediction module. The feature extraction module consists of a general feature extractor capable of handling a large number of visual categories. The density prediction module is designed to be related to the visual category. With training on common object dataset, FamNet presents fine adaption performances to novel object dataset by exporting Similarity Density Map (SDM) without fine-tuning, where the highlight areas in SDM indicate the possible locations of the target objects in query images (examples please refer to the highlight areas in SDM of Fig.~\ref{fig:frame_work}).

\vspace{-0.3cm} 
\subsection{Region Prediction using Region Proposal Network (RPN)}
The density map indicates the potential locations of novel objects. However, the possible regions of the novel objects are absent in SDM. To this end, we adopt Region Proposal Network (RPN) to generate potential bounding boxes for the novel objects. RPN is originally proposed in Faster-RCNN \cite{FasterRCNN2015}, used to generate class-agnostic proposals from the dense anchors.

\vspace{-0.3cm} 
\subsection{Object Identification}
The object identification consists of two steps: Class-specific Region Purification and Region Alignment. Class-specific Region Purification aims to remove the obvious partial false detections, while Region Alignment is used to align the regions. 

\vspace{-0.4cm} 
\subsubsection{Class-specific Region Purification}

SDM outputs the novel objects’ possible locations and RPN obtains region candidates in the query images. The regions indicated by RPN are class-agnostic proposals, while the Location Predictions presented by dot annotations in SDM are related to the object categories. Let $I_{SDM}^{q}$ be the SDM for a assigned query images, and $P_{F(j)}^{q}$ $(0\leq j\leq J)$ ($P_{F(j)}^{q}$ $\in$ $C_F^{q}$) is the $j^{th}$ candidate box in $C_{F}^{q}$, then the Class-specific Regions are given by (1).
\begin{small}
	\begin{equation}
	\frac{H(P_{F(j)}^{q})}{A(P_{F(j)}^{q})} \geq h
	\end{equation}
\end{small}

In (1), the function $H(.)$ is used to accumulate the highlighted intensity pixies in $I_{SDM}^{q}$ within the box given by $P_{F(j)}^{q}$. The intensity values in the dense map for each pixel are normalized to [0, 1). The function$ A(.)$ presents the total number of pixels in the region given by $P_{F(j)}^{q}$. If the value of (1) is true, an object is possibly contained in $P_{F(j)}^{q}$, where $h$ is a threshold. In this work, we follow the article \cite{Viresh2021}, and set its value as 0.1 empirically.

\vspace{-0.3cm} 
\subsubsection{Region Alignment Network (RAN)}
After the false detections are removed from the proposal boxes using Class-specific Region Purification, the left proposals are needed to judge whether they are category-related. Meanwhile, the category-related regions are necessarily aligned to the ground truth according to their central locations, widths, and heights. To this end, we propose a Region Alignment Network (RAN) to combine these two functionalities simultaneously.

The proposed RAN is construed on a Deep Siamese Networks (DSN) structure. Siamese networks consist of twin networks with sharing weights, where each network is respectively fed with a support image and a query one \cite{Gregory2015}. Early, siamese networks were usually used for object matching or image retrieving. Recently, deep siamese network (DSN) combined with deep feature extracted CNN) performs well in the visual object tracking tasks \cite{Luca2016,Boli2016}.

Inspired by the idea of DSN, we use it as the backbone of Region Alignment Network (RAN) in this work. One of the twin branches of RAN is connected to $C_{F(k)}^{s}$($1 \leq k \leq K$), where $C_{F(k)}^{s}$ is the $k^{th}$ shot images given by users. Another branch of DSN is connected to $\widetilde P_{F(\widetilde j)}^{q}$, where $\widetilde P_{F(\widetilde j)}^{q}$ $\in$ $P_{F(j)}^{q}$ ($0\leq \widetilde j \leq \widetilde J, \widetilde J \leq J$) is the $\widetilde j^{th}$ category-related region given by Class-specific Region Purification step.

The procedure of regression is presented as the Region Alignment Network (RAN) module in Fig.~\ref{fig:frame_work}. All image patches given by $C_{F(i)}^{s}$ and $\widetilde P_{F(\widetilde j)}^{q}$ are resized to W$\times$H (256$\times$256 in this work). We adopt a classical CNN structure ResNet50 \cite{Resnet50} as a branch to convert 256$\times$256 image to a feature vector with a length of $L$. Then we combine the two vectors obtained from the DSN two twin braches as a fused vector with a length of 2$L$. Then multiple layers of full connection neural network with a structure of 2$L$-$L$-$\frac{1}{2}$$L$-$\frac{1}{4}$$L$-16-5 is used to regress $2L$ vector to the target vector $\xi$($c$, $\vartriangle$x, $\vartriangle$y, $s_{w}$, $s_{h}$). In vector $\xi$, $c$ is calculated by (2), where function $\zeta$ indicates the category of $C_{F(i)}^{s}$ and $\widetilde P_{F(\widetilde j)}^{q}$. $c$ is used to suppress false categories. In this work, the value of $L$ is set to 1024.


\begin{small}
	\begin{equation}
	c=
	\begin{cases}
	1 \qquad if \quad \zeta(C_{F(i)}^{s}) == \zeta(\widetilde P_{F(\widetilde j)}^{q}) \\
	0 \qquad if \quad \zeta(C_{F(i)}^{s}) \neq \zeta(\widetilde P_{F(\widetilde j)}^{q})
	\end{cases}
	\end{equation}
\end{small}

$\vartriangle$x, $\vartriangle$y are the distance deviations between the predicted object center and the human annotated. They are calculated by (3), (4).

\vspace{-0.2cm} 
\begin{small}
	\begin{equation}
	\vartriangle x = 1 + \frac{2(x_{j}{'}-x_{i}{'})}{W}
	\end{equation}
\end{small}
\vspace{-0.2cm} 
\begin{small}
	\begin{equation}
	\vartriangle y = 1 + \frac{2(y_{j}{'}-y_{i}{'})}{H}
	\end{equation}
\end{small}

In (3) and (4), $x_{i}^{'}$=${x_i}\times \frac{w_{i}^{s}}{W}$, $y_{i}^{'}$=${y_i}\times \frac{w_{i}^{s}}{H}$ are the center point location of $C_{F(i)}^{s}$ in the normalized image patch, where $w_{i}^{s}$ and $h_{i}^{s}$ are the original width and height of $C_{F(i)}^{s}$ labeled by humans. Similarly, 
$x_{j}^{'}$=${x_j}\times \frac{w_{j}^{q}}{W}$, $y_{j}^{'}$=${y_j}\times \frac{w_{j}^{q}}{H}$ are the center point location of the $P_{F(j)}^{q}$ in the normalized image patch, where $w_{j}^{q}$ and $h_{j}^{q}$ are the predicted box width and height of $C_{F(i)}^{s}$ obtained by RPN. Using (3) and (4), the values of $\vartriangle$x, $\vartriangle$y are resized to the range of [0.0, 1.0) for regression.

$s_{w}$, $s_{h}$ are the scale difference along width and height between $C_{F(i)}^{s}$ and $P_{F(j)}^{q}$. They are calculated by (5), (6), where $\lambda_{w}$$\textgreater$1 and $\lambda_{h}$$\textgreater$1, indicate the maximal upscale times of $P_{F(j)}^{q}$ to $C_{F(i)}^{s}$. Conversely, $\frac{1}{\lambda_{w}}$, and $\frac{1}{\lambda_{h}}$ restrict the maximal downscale times of $P_{F(j)}^{q}$ to $C_{F(i)}^{s}$. In experiment, the values of $\lambda_{w}$ and $\lambda_{h}$ are set to 2.0. It means that the scale times between $P_{F(j)}^{q}$ and $C_{F(i)}^{s}$ along width and height are in the range of (0.5, 2.0). According to (5) and (6), the values of $s_{w}$, $s_{h}$ are resized to the range of (0.0, 1.0) for regression.

\vspace{-0.2cm} 
\begin{small}
	\begin{equation}
	s_{w} = \frac{\lambda_{w}\frac{w_{j}^{q}}{w_{i}^{s}}-1}{\lambda_{w}^{2}-1}
	\end{equation}
\end{small}
\vspace{-0.2cm} 
\begin{small}
	\begin{equation}
	s_{h} = \frac{\lambda_{h}\frac{h_{j}^{q}}{h_{i}^{s}}-1}{\lambda_{h}^{2}-1}
	\end{equation}
\end{small}


\section{Experiments}

In the the experiments, we first discuss the previous training of the proposed method before applying it to FSOD-AO. Then we compare the proposed method with the pioneer FSOD-AO ones on MS COCO \cite{COCO2014} and PASCAL VOC \cite{VOC2010} datasets.


\vspace{-0.2cm} 
\subsection{Pre-training for RAN before being Employed to FSOD-AO}


Following prior works \cite{AirDet2022,ARPN2020,FSOD-GCN2021}, the 80 classes in COCO are split into 60 non-VOC base classes and 20 novel classes. During training, the base class images from COCO trainval2014 are considered available. Let $P_{r}^{RPN}$ ($1\leq r \leq R$) be an image patch given by a region proposal and $P_{\widetilde r}^{GT}$ ($1\leq \widetilde r \leq \widetilde R, \widetilde R \leq R$) be the corresponding image patch which owns the ground truth region labeled by humans. Then in the training, the image pair $P_{r}^{RPN}$ and $P_{\widetilde r}^{GT}$ is similar to the pair of $C_{F(i)}^{s}$ and $\widetilde P_{F(\widetilde j)}^{q}$ (section ``Region Alignment Network"). The location distances between their centers, the scale differences between their width and height are calculated by (3)-(6), and used to guide the training of Region Alignment Network. There are quite a few cases in that no object (background) or only a very small part of objects are contained in the image patches in region proposals. In these cases, the objects in $P_{r}^{RPN}$ (or $\widetilde P_{F(\widetilde j)}^{q}$) do not belong to any categories of the objects in $P_{\widetilde r}^{GT}$ (or $C_{F(i)}^{s}$). These image pairs are also used in the training of RAN, where the parameter $c$ is set to 0. In this way, a total of 21018 image pairs are used to train RAN. The training of RAM is carried out on a computer with an Intel(R) Core(TM) i9-9900K 3.60GHz CPU, 32G memory, and V100 GPU card with 16G memory. And in the test phase, the models are deployed on a computer with 2.60GHz CPU, 8.0G RAM, and 2 NVIDIA Titan-X GPUs.


In the RAN test phase, 6728 pairs of $P_{r}^{RPN}$($x_{r}$, $y_{r}$, $w_{r}$, $h_{r}$) and $P_{\widetilde r}^{GT}$($\widetilde x_{r}^{gt}$, $\widetilde y_{r}^{gt}$, $\widetilde w_{r}^{gt}$, $\widetilde h_{r}^{gt}$) are used to analyze the performance of RAN for 20 novel class objects after it was trained on 60 base classes (21018 image pairs). Let $\widetilde P_{r}^{RAN}$($\widetilde x_{r}$, $\widetilde y_{r}$, $\widetilde w_{r}$, $\widetilde h_{r}$) be the aligned region for $P_{r}^{RPN}$ after it is processed by RAN, where $\widetilde x_{r}$, $\widetilde y_{r}$, $\widetilde w_{r}$, $\widetilde h_{r}$ are the aligned values for $x_{r}$, $y_{r}$, $w_{r}$, $h_{r}$ respectively. In these 6728 pairs, the categories of 5645 pairs are correctly classified.The classification accuracy of RAN is 83.90$\%$.  

\begin{figure}
	\begin{center}
	\includegraphics[width=7.5cm]{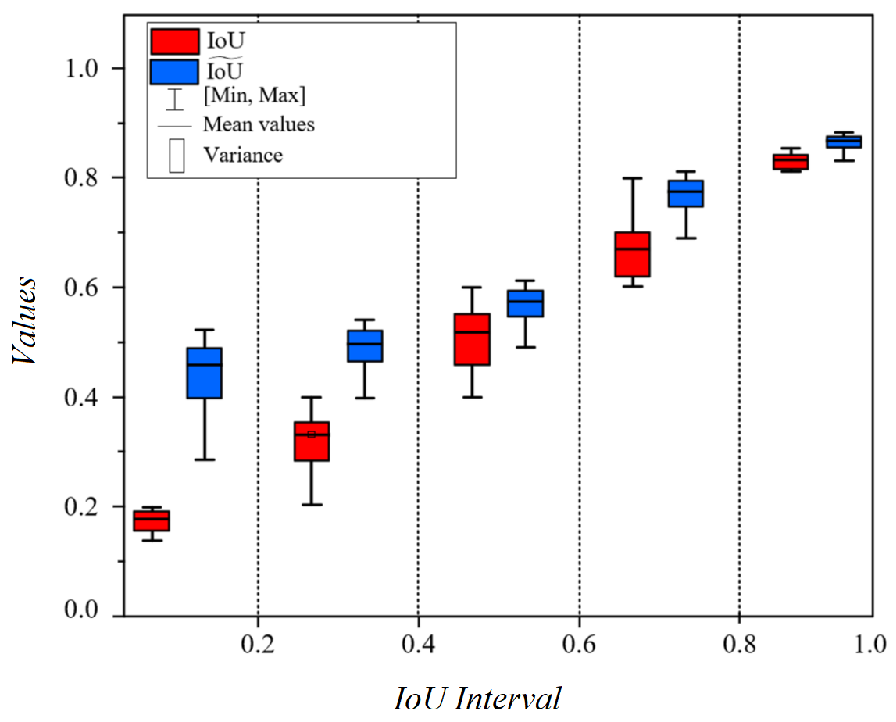}
	\end{center}
	\caption{The comparisons of the mean values, maximal values, minimum values and variance range between $IoU_{RPN}$ and $\widetilde{IoU}_{RAN}$ every 0.2 intervals.}
	\label{fig:RAN-IOU}
\end{figure}

We continuously discuss the performance of the RAN on region alignment. Fig. 2 presents the comparisons of the mean values, maximal values, minimum values, and variance range between $IoU_{RPN}$ and $\widetilde{IoU}_{RAN}$ every 0.2 intervals, where $IoU_{RPN}$ and $\widetilde{IoU}_{RAN}$ are the IoU (Intersection over Union) values of ($P_{r}^{RPN}$, $P_{\widetilde r}^{GT}$) and ($\widetilde P_{r}^{RAN}$, $P_{\widetilde r}^{GT}$) respectively. $IoU_{RPN}$ and $\widetilde{IoU}_{RAN}$ are given by (7) and (8), where function $A(.)$ indicates the area of the range. The higher the values of $\widetilde{IoU}_{RAN}$, the better performance for RAN. We can see from Fig. 2 that with the process of RAN, the aligned $\widetilde{IoU}_{RAN}$ obviously increased especially for the initial $IoU_{RPN}$ whose values in the range of [0.0, 0.2), [0.2, 0.4), [0.6, 0.8). The increments of IoU are about 0.28, 0.19, and 0.14 respectively. Meanwhile, the IoU variances for the ranges of [0.0, 0.2), [0.2, 0.4), [0.4, 0.6), [0.6, 0.8) obviously shrink. It indicates that the proposed RAN can effectively align the predicted objects’ region proposals (the RPN boxes) to the ground truth.

\begin{small}
	\begin{equation}
	IoU_{RPN} = \frac{A(P_{r}^{RPN} \cap P_{\widetilde r}^{GT})}{A(P_{r}^{RPN} \cup P_{\widetilde r}^{GT})}
	\end{equation}
\end{small}
\begin{small}
	\begin{equation}
	\widetilde{IoU}_{RAN} =  \frac{A(\widetilde P_{r}^{RAN} \cap P_{\widetilde r}^{GT})}{A(\widetilde P_{r}^{RAN} \cup P_{\widetilde r}^{GT})}
	\end{equation}
\end{small}

\begin{figure}
	\begin{center}
		\includegraphics[width=8cm]{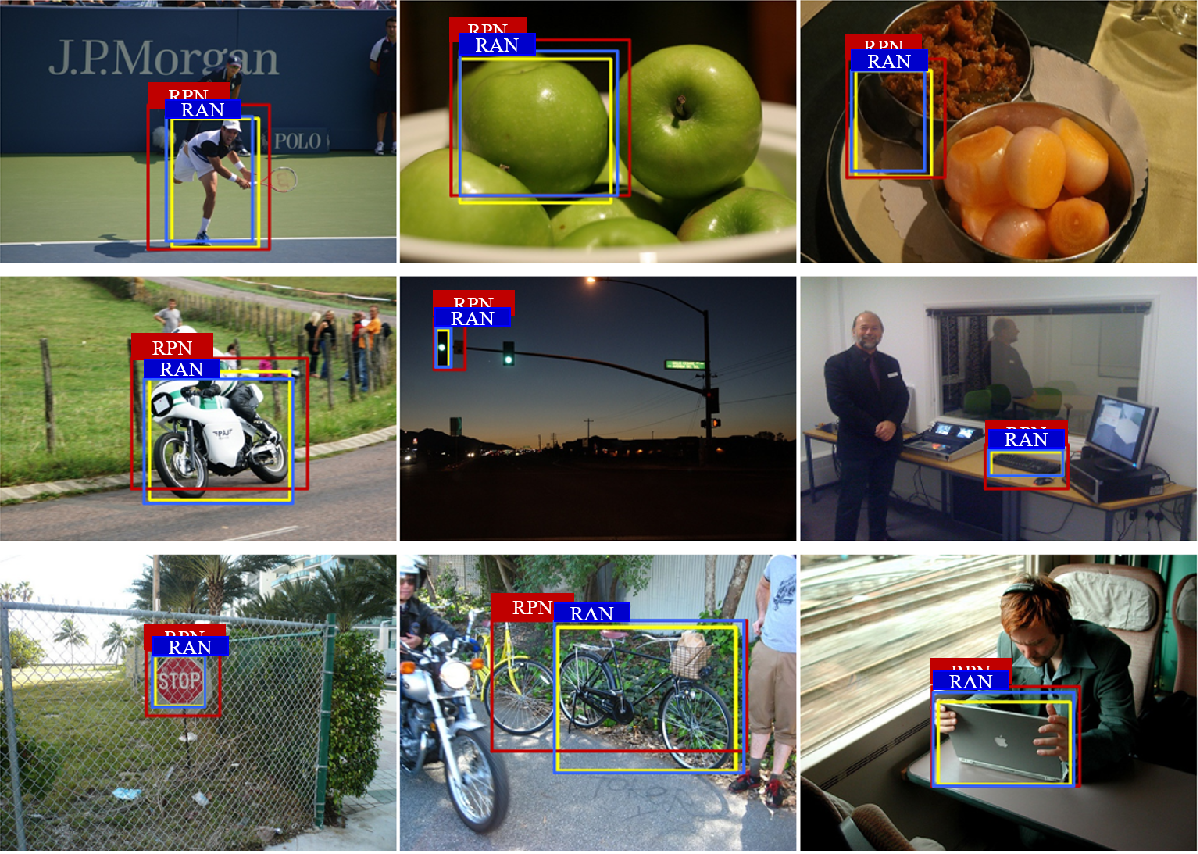}
	\end{center}
	\caption{Some COCO images' visualization results before and after RAN process, where the red, blue and yellow boxes are the regions of $P_{r}^{RPN}$, $\widetilde P_{r}^{RAN}$, and $P_{\widetilde r}^{GT}$ respectively.}
	\label{fig:RAN_Vis}
\end{figure}

Fig.~\ref{fig:RAN_Vis} shows the visualisation results of some COCO images before and after the RAN process, where the red, blue and yellow boxes indicate the regions of $P_{r}^{RPN}$, $\widetilde P_{r}^{RAN}$, $P_{\widetilde r}^{GT}$  respectively. From Fig.~\ref{fig:RAN_Vis}, we can see that RAN helps to align the RPN proposals with the ground truth, effectively improving the performance of the IoU.

\begin{table*}[!htbp]
	\centering
	\caption{Performance comparison on 20 novel categories COCO validation dataset. In each setting, red and green fonts denote the best and second-best performance for the methods. The underlined numbers represent the highest scores achieved by all methods, both with and without fine-tuning.}
	\begin{tabular}{c c c c c c c c c c c c c c }
		\hline
		\multicolumn{2}{c}{Shots} & \multicolumn{3}{c}{1} & \multicolumn{3}{c}{2} & \multicolumn{3}{c}{3} & \multicolumn{3}{c}{5}\\
		\cline{1-14}   
		Method & $\makecell{Fine\\Tune$?$}$ & $AP$  & $AP_{50}$ & $AP_{75}$ & $AP$  & $AP_{50}$ & $AP_{75}$ & $AP$  & $AP_{50}$ & $AP_{75}$ & $AP$  & $AP_{50}$ & $AP_{75}$ \\
		\hline
		\makecell{$TFA_{cos}$ \\ \cite{Wang2020}} & $\surd$ & 3.09 &5.24 &3.21 &4.21 &7.70 &4.35 &6.05 &11.48 &5.93 &7.61 &14.56 &7.17 \\
		
		\makecell{$FSDet$ \\ \cite{Yang2020}} & $\surd$ & 2.20 &6.20 &0.90 &3.40 &10.00 &1.50 &5.20 &14.70 &2.10 &8.20 &\textcolor{green}{21.60} &4.70 \\
		\makecell{A-RPN \\ \cite{ARPN2020}} & $\surd$ & 4.59 &8.85 &4.37 &6.15 &12.05 &5.76 &8.24 &15.52 &7.92 &9.02 &17.29 &8.53 \\
		
		\makecell{FSOD-GCN \\ \cite{FSOD-GCN2021}} & $\surd$  & - &- &- & 7.6 &16.1 &6.2 &- &- &- &- &- &- \\
		
		\makecell{AirDet \\ \cite{AirDet2022}} & $\surd$ & \textcolor{green}{6.10} &\textcolor{green}{11.40} &\textcolor{green}{6.04} &\textcolor{green}{8.73} &\textcolor{green}{16.24} &\textcolor{green}{8.35} &\textcolor{green}{9.95} &\textcolor{green}{19.39} &\textcolor{green}{9.09} &\textcolor{green}{10.81}& 20.75 &\underline{\textcolor{red}{10.27}} \\
		
		\makecell{FSOD-VFA \\ \cite{FSOD-VFA2023}} & $\surd$  & - &- &- & - &- &- &- &- &- &- &16.2 &- \\
		\makecell{DiGeo \\ \cite{DiGeo2023}} & $\surd$  & - &- &- & - &- &- &- &- &- &10.3 &18.7 &9.9 \\
		
		\makecell{A-RPN \\ \cite{ARPN2020}}  & $\times$ 	& 4.32 &7.62 &4.30 &4.67 &8.83 &4.49 &5.28 &9.95 &5.05 &6.08 &11.17 &5.88	 \\
		
		\makecell{AirDet \\ \cite{AirDet2022}} & $\times$ & 5.97 &10.52 &5.98 &6.58 &12.02 &6.33 &7.00 &12.95 &6.71 &7.76 &14.28 &7.31 \\
		
		\small{\makecell{SDM-RAN\\(Proposed)}} & $\times$ & \underline{\textcolor{red}{6.57}} &\underline{\textcolor{red}{14.30}} &\underline{\textcolor{red}{6.11}} &\underline{\textcolor{red}{8.87}} &\underline{\textcolor{red}{16.54}} &\underline{\textcolor{red}{8.58}} &\underline{\textcolor{red}{10.39}} &\underline{\textcolor{red}{20.65}} &\underline{\textcolor{red}{9.52}} &\underline{\textcolor{red}{11.05}} &\underline{\textcolor{red}{21.76}} &\textcolor{green}{10.24} \\
		\hline							
	\end{tabular}
	\begin{flushleft}
	\small{$^*$Numbers in the table are prediction accuracy and measurement is percentage ($\%$)}	
    \end{flushleft}
	\label{tab1}	
\end{table*}

\begin{table*}[!htbp]
	\centering
	\caption{Performance comparison of $AP_{50}$ on the PASCAL VOC dataset. In each setting, red and green fonts denote the best and second-best performance for the methods without fine-tuning. The underlined numbers represent the highest scores achieved by all methods, both with and without fine-tuning. The proposed SDM-RAN achieves better performance than other SOTAs without fine-tuning and yields competitive results compared to the methods with fine-tuning.}
	\begin{tabular}{c c c c c c c c c c c c c c }
		\hline
		\multicolumn{2}{c}{Novel Set} & \multicolumn{4}{c}{Split1} & \multicolumn{4}{c}{Split2} & \multicolumn{4}{c}{Split3} \\
		\cline{1-14}   
		Method & $\makecell{Fine\\Tune$?$}$ & 1  & 2 & 3 & 5  & 1 & 2 & 3  & 5 & 1 & 2  & 3 & 5 \\
		\hline
		\makecell{TFA \\ \cite{Wang2020}} & $\surd$ & \underline{36.8} & 29.1 & \underline{43.6} & \underline{55.7} & 18.2 & \underline{29.0} & \underline{33.4} & \underline{35.5} & \underline{27.7} & \underline{33.6} & \underline{42.5} & \underline{48.7} \\
		
		
		
		
		\makecell{A-RPN \\ \cite{ARPN2020}}  &$\surd$ & 17.41 & 23.66 & 26.30 & 28.08 & 12.19 & 18.24 & 21.17 & 23.25 & 14.91 & 21.66 & 26.3 & 30.08	 \\
		
		\makecell{AirDet \\ \cite{AirDet2022}} & $\surd$ & 24.64 & \underline{30.35} & 32.05 & 33.02 & 18.64 & 25.16 & 27.22 & 27.38 & 17.39 & 22.19 & 29.87 & 31.45 \\

		\makecell{A-RPN \\ \cite{ARPN2020}}  & $\times$ & 18.10 & 22.6 & 24.08 & 25.03 & 13.34 & 19.26 & 21.17 & 22.93 & 16.58 & 20.14 & 22.26 & 23.77	 \\
		
		\makecell{AirDet \\ \cite{AirDet2022}} & $\times$ & \textcolor{green}{21.33} & \textcolor{green}{26.80} & \textcolor{green}{28.61} & \textcolor{green}{29.78} & \textcolor{green}{16.18} & \textcolor{red}{21.48} & \textcolor{red}{23.55} & \textcolor{green}{24.29} & \textcolor{green}{22.18} & \textcolor{red}{24.23} & \textcolor{green}{26.54} & \textcolor{green}{28.37} \\
		
		\small{\makecell{SDM-RAN\\(Proposed)}} & $\times$ & \textcolor{red}{24.65} & \textcolor{red}{27.79} & \textcolor{red}{29.42} & \textcolor{red}{31.58} & \underline{\textcolor{red}{19.06}} & \textcolor{green}{20.43} & \textcolor{green}{23.17} & \textcolor{red}{27.15} & \textcolor{red}{23.3} & \textcolor{green}{24.21} & \textcolor{red}{27.60} & \textcolor{red}{29.12} \\
		\hline							
	\end{tabular}
	\begin{flushleft}
	\small{$^*$Numbers in the table are prediction accuracy and measurement is percentage ($\%$)}
	\end{flushleft}
	\label{tab1}	
\end{table*}

\begin{table*}
	\caption{Effciency comparison with offcial source code. The red and green fonts denote the best and second-best performance.}	
	\centering
	\begin{tabular}{c c c c c c c }
		\hline
		
		Method  & \makecell{SDM-RAN \\ (Proposed)} & \makecell{AirDet \\\cite{AirDet2022}} & \makecell{A-RPN \\ \cite{ARPN2020}} & \makecell{FSDet \\\cite{Yang2020}} & \makecell{MPSR \\ \cite{MPSR2020}} & \makecell{$TFA_{cos}$ \\ \cite{Wang2020}}   \\ 
		\hline				
		\makecell{Fine-tuning \\ (min)} & 0 & 0 & 0 & 11 & 3 & -   \\ 
		
		\makecell{Inference \\ (s/img)} & \textcolor{red}{0.043} & 0.081 & \textcolor{green}{0.076} & 0.109 & 0.202 & 0.94   \\ 
		\hline	
		
	\end{tabular}
	\label{tab1}

\end{table*}

\begin{table*}[!htbp]
	\caption{Effect of different modules of AP scores on COCO. Red and green fonts are used in each group to indicate the highest and second-highest performance improvement for the module.}	
	\centering
	\begin{tabular}{c c c c c c c c c c c c c c c }
		\hline
		\multicolumn{3}{c}{Shots} & \multicolumn{3}{c}{1} & \multicolumn{3}{c}{2} & \multicolumn{3}{c}{3} & \multicolumn{3}{c}{5}\\
		\cline{1-15}   
		SDM & RPN & RAN & $AP$  & $AP_{50}$ & $AP_{75}$ & $AP$  & $AP_{50}$ & $AP_{75}$ & $AP$  & $AP_{50}$ & $AP_{75}$ & $AP$  & $AP_{50}$ & $AP_{75}$ \\
		\hline
		$\surd$ & - & - & 2.76 & 5.78 & 2.12 & 3.68 & 6.39 & 2.31 & 3.88 & 8.34 & 3.45 & 5.32 & 9.72 & 4.87   \\
		$\surd$ & $\surd$ & - &  4.52 & 12.39 & 3.01 & 5.27 & 12.45 & 4.79 & 7.22 & 17.99 & 5.56 & 7.98 & 19.05 & 6.79  \\
		$ - $ & $+\Delta$ & - & $\textcolor{green}{1.76}$ & $\textcolor{red}{6.61}$  & 0.89 & 1.59 & $\textcolor{red}{6.06}$ & $\textcolor{green}{2.48}$ & $\textcolor{green}{3.34}$ & $\textcolor{red}{9.65}$ & 2.11 & $\textcolor{green}{2.66}$ & $\textcolor{red}{9.33}$ & 1.92  \\
		$\surd$ & $\surd$ & $\surd$ & 6.57 & 14.3 & 6.11 & 8.87 & 16.54 & 8.58 & 10.39 & 20.65 & 9.52 & 11.05 & 21.76 & 10.54   \\
		$  -  $ & $  -  $ & $+\Delta$ & $\textcolor{green}{2.05}$ & 1.91 & $\textcolor{red}{3.1}$ & 3.6 & $\textcolor{red}{4.09}$ & $\textcolor{green}{3.79}$ & $\textcolor{green}{3.17}$ & 1.66 & $\textcolor{red}{3.96}$ & $\textcolor{green}{3.07}$ & 1.71 & $\textcolor{red}{3.75}$   \\
		\hline				
		
	\end{tabular}	
    \begin{flushleft}
    \small{$^*$Numbers in the table are prediction accuracy and measurement is percentage ($\%$)}
    \end{flushleft}
	\label{tab1}

\end{table*}


\vspace{-0.2cm} 
\subsection{Comparison with State-of-the-Art FSOD-AO methods}

\vspace{-0.2cm} 
\subsubsection{Coco Dataset}

We first present the evaluation on COCO benchmark \cite{COCO2014}, where the model trained and tested on COCO dataset. After the pre-training of the model on the 60 classes of trainval2014, we evaluate the model on 5,000 images from 20 novel classes in COCO without fine-tuning. Similarly, we adopt the typical metrics \cite{AirDet2022}, i.e. $AP$, $AP_{50}$ and $AP_{75}$ for evaluation. the supported images for 20 novel classes in COCO

As shown in Table 1, our method achieves a significant performance gain than those of A-RPN \cite{ARPN2020} and AirDet \cite{AirDet2022} without fine-tuning, especially for the values of $AP_{50}$. And our method also achieves comparable or even better results than the fine-tuning methods, such as FSOD-VFA \cite{FSOD-VFA2023}, DiGeo \cite{DiGeo2023}, FSOD-GCN \cite{FSOD-GCN2021}, $TFA_{cos}$ \cite{Wang2020}, and FSDet \cite{Yang2020}. Meanwhile, we can see from Table 1 that the values of $AP_{50}$ of 1-shot are even better than those of AirDet without fine-tuning under 2, 3, 5 shot situations. It is because the SDM step can effectively outline similar areas in the query images for the given support images, and RAN can effectively align the candidates to the ground truth. 

\vspace{-0.4cm} 
\subsubsection{PASCAL VOC Dataset}

Similar to the work \cite{Wang2020}, for the few-shot PASCAL VOC dataset, the 20 classes are randomly divided into 15 base classes and 5 novel classes, where the novel classes have K = 1, 2, 3, 5 objects per class sampled from the test set of VOC2007. Three random split groups are considered in this work and $AP_{50}$ (matching threshold is 0.5) of the novel classes is used on PASCAL VOC. Table 2 presents a performance comparison of $AP_{50}$ between our method and SOTAs on the PASCAL VOC dataset. The numbers of state-of-the-art (SOTA) results listed in Table 2 were obtained directly from official source code or extracted from the relevant articles.

Our method also performs well on PASCAL VOC. As shown in Table 2, the proposed method achieves the best (9/12) or second-best (3/12) among all FSOD methods without fine-tuning. Especially when the shot is low, our method shows obvious improvements. For example, our 1-shot gains are 3.32$\%$, 2.88$\%$ and 1.12$\%$ on the novel set 1, 2 and 3 respectively compared to the second-best FSOD ones without fine-tuning. In addition, our method also yields competitive results compared to the SOTA FSOD methods with fine-tuning.

\subsection{Efficiency Comparison}
Table 3 shows the efficiency comparison between SDM-RAN and the SOTA methods with the official source code. Similar to the SOTA methods, SDM-RAN also runs at a setting of 3 shots per class using 2 NVIDIA Titan-X GPUs (AirDet was employed on 4 NVIDIA Titan-X GPUs). Without fine-tuning, SDM-RAN and AirDet can make direct inferences on novel objects with comparable speed, while the other methods require a fine-tuning time of about 3-11 minutes. Dislike the relatively complex multi-relation detector in A-PRN and multiple time-consuming modules (Global-Local Relation, Embedded Location Regression) in AirDet, the proposed method consists of three straightforward phases, including SDM (a lightweight convolution procedure), PRN (region prediction), and RAN (region alignment). Each module runs independently in less than 0.013 seconds per frame. Therefore, SDM-RAN operates at a speed of 0.043 seconds per frame, which is significantly faster than AirDet and A-RPN.

\vspace{-0.2cm} 
\subsection{Ablation Studies}
We conduct a series of ablation experiments on COCO dataset. 

\vspace{-0.4cm} 
\subsubsection{How effective is PRN adopted on basic SDM module?}
Table 4 lists the $AP$, $AP_{50}$, $AP_{75}$ values for basic SDM module, SDM(PRN) and SDM(PRN)-RAN in detail. The SDM module originally provides only location information, without any indication of the regions. Therefore we adopt the size of the objects in support images as the default anchor for the targets generated by SDM. From the values listed in the $5^{th}$ row of Table 4, we can see that PRN significantly improve the $AP_{50}$ values from the basic SDM module at high value of 9.65$\%$. In addition, the use of PRN also clearly enhances $AP$ scores from SDM.

\vspace{-0.4cm} 
\subsubsection{How effective is RAN adopted to SDM(RPN)?}
We continually analyze the performance of RAN after its adoption for the combination of SDM and RPN. The values in the $7^{th}$ row of Table 4 indicate that RAN contributes significantly to improving $AP_{75}$ scores, with an increase of between 3.1$\%$ and 3.96$\%$ compared to SDM(RPN). Furthermore, RAN enhances the $AP$ values compared to SDM(RPN), with a low increase of 2.05$\%$ and a high increase of 3.6$\%$.

\vspace{-0.4cm} 
\subsubsection{Contributions of RAN}
Table 4 shows that RPN can significantly improve the $AP_{50}$ values of the basic SDM by combining preliminary target regions with SDM locations, depending on the position identification function of SDM. Furthermore, based on the preliminary RPN regions, RAN can continually refine the targets' ranges using the proposed deep siamese structure. For instance, the first, fifth, and eighth images in Fig. 3 show the superior region refining performance of the blue regions (generated by RAN) compared to the red ones (outputted by RPN) in detecting the targets person, lamp, and bicycle. This demonstrates the effectiveness of RAN in aligning with the ground truths. Therefore, RAN primarily contributes to an improvement in the $AP_{75}$ score compared to SDM(RPN), with an additional increase in the $AP$.

\vspace{-0.2cm} 
\subsection{Discussions}
\vspace{-0.2cm} 
The proposed FSOD-AO method is tested experimentally and compared with the state-of-the-art methods on COCO and VOC datasets. In general, the results on COCO and VOC datasets clearly show that the proposed method outperforms A-RPN and AirDet with 1, 2, 3 and 5 shots on $AP_{50}$. Partially, compared to the two pioneering works (A-RPN and AirDet), the proposed method gains higher $AP_{75}$ scores with 1, 2 and 3 shots on COCO, and obtain better $AP$ values with 1, 3, 5 shots on VOC. In addition, the proposed method also archives competitive results on $AP_{50}$ compared to the SOTA methods with fine tuning, such as FSOD-VFA \cite{FSOD-VFA2023}, DiGeo \cite{DiGeo2023}, FSOD-GCN \cite{FSOD-GCN2021}, TFA \cite{Wang2020} on COCO dataset. Our proposed method provides a novel solution for the FSOD-AO task by simultaneously evaluating both the object locations and region proposals. It is noticeable that when selecting one or two support images for novel classes, it is recommended to choose the most commonly seen ones as supported images. However, for 3, 5 or more shots, randomly selected images work well.

It is noticeable that we set the values of $h$, $\lambda_{w}$ and $\lambda_{h}$ empirically to 0.1, 2.0. The threshold $h$ is used to determine whether a highlighted intensity region in SDM could be selected as a candidate. And $\lambda_{w}$ and $\lambda_{h}$ are used to limit the maximum and minimum size scale for the objects in the query images to those annotated in the support images. In practical object detection tasks, such as autonomous exploration in unknown environments, fast counting of dense objects, etc., these parameters can be set dynamically according to the size of the targets in the query images.


\vspace{-0.3cm} 
\section{Conclusions}
\vspace{-0.2cm} 
This work presents a novel technical realization for the task of correctly identifying the categories of new objects, while simultaneously extracting their precise regions without fine-tuning. We first adopt an attention-based structure (SDM) to locate the novel object. Additionally, we incorporate RPN into our framework to obtain object regions. The combination of SDM and RPN enables the natural generation of approximate positions and regions for novel objects. To accurately match and align the approximate candidates with the supported images (patches), we propose a Deep Siamese Networks based structure (RAN). This structure serves to eliminate false candidates and synchronously align the region by analyzing the differences in location and range scale between the predicted and ground truth boxes. With the pre-learning on the annotated labels given by traditional datasets, the proposed SDM-RAN can identify a previously unseen object immediately without fine tuning. Experiments are conducted on the MS COCO and PASCAL VOC datasets, and the results show that the proposed method outperforms the SOTAs on the FSOD-AO task. Furthermore, it offers the advantage of faster execution compared to traditional FSOD-AO methods.

\section{Acknowledgements}
This work is supported by the National Key Research and Development Program of China (No. 2018AAA0102902), the Guangxi Key Research and Development Program (AB24010164, AB21220038, AB23026048), the National Natural Science Foundation of China (NSFC) (No.61873269), the Beijing Natural Science Foundation (J210012, L192005), the Hebei Natural Science Foundation (F2021205014).

\vspace{-0.2cm} 
\bibliographystyle{splncs04}
\bibliography{egbib.bib}

\begin{thebibliography}{10}
\providecommand{\url}[1]{\texttt{#1}}
\providecommand{\urlprefix}{URL }
\providecommand{\doi}[1]{https://doi.org/#1}

\bibitem{Aming2021}
Aming~Wu, Yahong~Han, L.Z.Y.Y.: Universal-prototype enhancing for few-shot object detection. In: Proceedings of the IEEE/CVF International Conference on Computer Vision (ICCV 2021). pp. 9567--9576 (2021)

\bibitem{Andriy2010}
Andriy~Myronenko, X.S.: Point set registration: Coherent point drift. IEEE Transactions on Pattern Analysis and Machine Intelligence  \textbf{32}(12),  2262--2275 (2010)

\bibitem{Bharath2017}
Bharath~Hariharan, R.G.: Low-shot visual recognition by shrinking, hallucinating features. In: IEEE International Conference on Computer Vision (ICCV 2017). Venice, Italy (October 2017)

\bibitem{Boli2016}
Bo~Li, Junjie~Yan, W.W.Z.Z.X.H.: High performance visual tracking with siamese region proposal network. In: IEEE/CVF Conference on Computer Vision, Pattern Recognition (CVPR 2018). Salt Lake City, UT, USA (June 2018)

\bibitem{Sunbo2021}
Bo~Sun, Banghuai~Li, S.C.Y.Y.C.Z.: Fsce: Few-shot object detection via contrastive proposal encoding. In: Proceedings of the IEEE/CVF Conference on Computer Vision, Pattern Recognition (CVPR 2021). pp. 7352--7362. Nashville, TN, USA (June 2021)

\bibitem{Boris2018}
Boris N.~Oreshkin, Pau~Rodriguez, A.L.: Tadam: Task dependent adaptive metric for improved few-shot learning. In: Proceedings of the 31st International Conference on Neural Information Processing Systems (NIPS 2018). p. 719¨C729 (December 2018)

\bibitem{AirDet2022}
Bowen~Li, Chen~Wang, P.R.S.K.S.S.: Airdet: Few-shot detection without fine-tuning for autonomous exploration. In: Proceedings of the European Conference on Computer Vision (ECCV 2022). Tel-Aviv, Israel (October 2022)

\bibitem{Carlos2016}
Carlos~Arteta, Victor~Lempitsky, r.Z.: Counting in the wild. In: Proceedings of the European Conference on Computer Vision (ECCV 2016). Amsterdam, The Netherlands (October 2016)

\bibitem{Chenchen2021}
Chenchen~Zhu, Fangyi~Chen, U.A.Z.S.M.S.: Semantic relation reasoning for shot-stable few-shot object detection. In: Proceedings of the IEEE/CVF Conference on Computer Vision, Pattern Recognition (CVPR 2021). Nashville, TN, USA (June 2021)

\bibitem{Eleni2017}
Eleni~Triantafillou, Richard~Zemel, R.U.: Few-shot learning through an information retrieval lens. In: Proceedings of the 31st International Conference on Neural Information Processing Systems (NIPS 2017). pp. 2252 -- 2262 (December 2017)

\bibitem{Pedro2005}
Felzenszwalb, P.F.: Representation and detection of deformable shapes. IEEE Transactions on Pattern Analysis and Machine Intelligence  \textbf{39}(7),  208 -- 220 (2005)

\bibitem{Finn2017}
Finn~Chelsea, Abbeel~Pieter, L.S.: Model-agnostic meta-learning for fast adaptation of deep networks. In: International Conference on Machine Learning (ICML 2017). Williamstown, MA, USA (July 2017)

\bibitem{Gregory2015}
Gregory~Koch, Richard~Zemel, R.S.: Siamese neural networks for one-shot image recognition. In: 2015 International Conference on Machine Learning Deep Learning workshop (2015)

\bibitem{FSOD-GCN2021}
Guangxing~Han, Yicheng~He, S.H.J.M.S.F.C.: Query adaptive few-shot object detection with heterogeneous graph convolutional networks. In: IEEE International Conference on Computer Vision (ICCV 2021). Montreal, Canada (Oct 2021)

\bibitem{Jake2017}
Jake~Snell, Kevin~Swersky, R.Z.: Prototypical networks for few-shot learning. In: Proceedings of the 31st International Conference on Neural Information Processing Systems (NIPS 2017). p. 4080¨C4090 (December 2017)

\bibitem{FSOD-VFA2023}
Jiaming~Han, Yuqiang~Ren, J.D.K.Y.G.S.X.: Few-shot object detection via variational feature aggregation. In: Proceedings of the37th AAAI Conference on Artificial Intelligence (AAAI 2023). Washington DC, USA (Feb 7 2023)

\bibitem{DiGeo2023}
Jiawei~Ma, Yulei~Niu, J.X.S.H.G.H.S.F.C.: Digeo: Discriminative geometry-aware learning for generalized few-shot object detection. In: Proceedings of the IEEE / CVF Computer Vision, Pattern Recognition Conference (CVPR) (CVPR 2023). Vancouver, Canada (June 2023)

\bibitem{MPSR2020}
Jiaxi~Wu, Songtao~Liu, D.H.Y.W.: Multi-scale positive sample refinement for few-shot object detection. Proceedings of the European Conference on Computer Vision (ECCV) pp. 456 -- 472 (2020)

\bibitem{Resnet50}
Kaiming~He, Xiangyu~Zhang, S.R.J.S.: Deep residual learning for image recognition. In: IEEE Conference on Computer Vision, Pattern Recognition (CVPR 2016). Las Vegas, NV, USA (June 2016)

\bibitem{Kwonjoon2019}
Kwonjoon~Lee, Subhransu~Maji, A.R.S.S.: Meta-learning with differentiable convex optimization. In: IEEE Conference on Computer Vision, Pattern Recognition (CVPR 2019). pp. 10657 -- 10665. Long Beach, CA, USA (June 2019)

\bibitem{Laura2021}
Laura~Mannocci, S¨¦bastien~Villon, M.C.N.G.N.M.C.I.L.V.D.M.: Leveraging social media, deep learning to detect rare megafauna in video surveys. Conservation Biology  \textbf{36}(1) (June 2021)

\bibitem{WeiLian2012}
Lian, W.: Rotation-invariant nonrigid point set matching in cluttered scenes. IEEE Transactions on Image Processing  \textbf{21}(5),  2786 -- 2797 (2012)

\bibitem{Luca2016}
Luca~Bertinett, Jack~Valmadre, J.F.H.r.V.P.H.S.T.: Fully-convolutional siamese networks for object tracking. In: European Conference on Computer Vision (ECCV 2016). pp. 10657 -- 10665. Long Beach, CA, USA (June 2019)

\bibitem{VOC2010}
Mark~Everingham, Luc Van~Gool, C.K.W.J., Zisserman, A.: The pascal visual object classes (voc) challenge. International journal of computer vision  \textbf{88}(2),  303 -- 338 (June 2010)

\bibitem{Oriol2016}
Oriol~Vinyals, Charles~Blundell, T.L.K.K.D.W.: Matching networks for one shot learning. In: Proceedings of Advances in Neural Information Processing Systems (NIPS 2016). p. 3630¨C3638 (2016)

\bibitem{ARPN2020}
Qi~Fan, Wei~Zhuo, C.K.T.Y.W.T.: Few-shot object detection with attention-rpn, multi-relation detector. In: Proceedings of IEEE/CVF Conference on Computer Vision, Pattern Recognition (CVPR 2020). Seattle, WA, USA (June 2020)

\bibitem{FasterRCNN2015}
Shaoqing~Ren, Kaiming~He, R.G.J.S.: Faster r-cnn: Towards real-time object detection with region proposal networks. In: Neural Information Processing Systems (NIPS 2015) (2015)

\bibitem{Timothy2022}
Timothy~Hospedales, Antreas~Antoniou, P.M.A.S.: Meta-learning in neural networks: A survey. IEEE Transactions on Pattern Analysis, Machine Intelligence (TPAMI)  \textbf{44}(9),  5149 -- 5169 (2022)

\bibitem{Tsendsuren2018}
Tsendsuren~Munkhdalai, Xingdi~Yuan, S.M.A.T.: Rapid adaptation with conditionally shifted neurons. In: International Conference on Machine Learning (ICML 2018). pp. 2252 -- 2262. Stockholm, Sweden (July 2018)

\bibitem{COCO2014}
Tsung-Yi~Lin, Michael~Maire, S.B.J.H.P.P.D.R.P.D.C.L.Z.: Microsoft coco: Common objects in context. In: Proceedings of the European Conference on Computer Vision (ECCV 2014). pp. 740 -- 755 (2014)

\bibitem{Viresh2021}
Viresh~Ranjan, Udbhav~Sharma, T.N.M.H.: Learning to count everything. In: Proceedings of the IEEE conference on computer vision, pattern recognition (CVPR 2021). Nashville, TN, USA (June 2021)

\bibitem{Wang2020}
Xin~Wang, Thomas E.~Huang, T.D.J.E.G.F.Y.: Frustratingly simple few-shot object detection. In: Proceedings of the International Conference on Machine Learning (ICML 2020). pp. 1 -- 12 (2020)

\bibitem{Xun2019}
Xun~Wang, Xintong~Han, W.H.D.D.M.R.S.: Multi-similarity loss with general pair weighting for deep metric learning. In: IEEE Conference on Computer Vision, Pattern Recognition (CVPR 2019). Long Beach, CA, USA (June 2019)

\bibitem{Yang2020}
Yang~Xiao, Vincent~Lepetit, R.M.: Few-shot object detection, viewpoint estimation for objects in the wild. In: Proceedings of the International Conference on Machine Learning (ICML 2020). pp. 1--12 (2020)

\end{thebibliography}


\end{document}